\documentclass[10pt,twocolumn,letterpaper]{article}

\usepackage{cvpr}
\usepackage{times}
\usepackage{epsfig}
\usepackage{graphicx}
\usepackage{amsmath}
\usepackage{amssymb}
\usepackage{enumerate}
\usepackage{mathrsfs}
\usepackage{bm}
\usepackage[ruled,vlined,lined,boxed,linesnumbered]{algorithm2e}
\usepackage{subfigure}
\graphicspath{{figures/}}

\usepackage[pagebackref=true,breaklinks=true,letterpaper=true,colorlinks,bookmarks=false]{hyperref}

\cvprfinalcopy 

\def\cvprPaperID{3636} 

\ifcvprfinal\fi
\begin{document}

\title{Multispectral Image Intrinsic Decomposition via Low Rank Constraint}

\author{Qian Huang\\
Nanjing University\\
{\tt\small cianhwang@gmail.com}
\and
Weixin Zhu\\
Nanjing University\\
{\tt\small njuzhwx@163.com}
\and
Yang Zhao\\
Nanjing University\\
{\tt\small zhaoyang@smail.nju.edu.cn}
\and
Linsen Chen\\
Nanjing University\\
{\tt\small njucls@163.com}
\and
Yao Wang\\
New York University\\
{\tt\small yw523@nyu.edu}
\and
Tao Yue\\
Nanjing University\\
{\tt\small yuetao@nju.edu.cn}
\and
Xun Cao\\
Nanjing University\\
{\tt\small caoxun@nju.edu.cn}
}

\maketitle

\begin{abstract}
Multispectral images contain many clues of surface characteristics of the objects, thus can be widely used in many computer vision tasks, e.g., recolorization and segmentation. However, due to the complex illumination and the geometry structure of natural scenes, the spectra curves of a same surface can look very different. In this paper, a Low Rank Multispectral Image Intrinsic Decomposition model (LRIID) is presented to decompose the shading and reflectance from a single multispectral image. We extend the Retinex model, which is proposed for RGB image intrinsic decomposition, for multispectral domain. Based on this, a low rank constraint is proposed to reduce the ill-posedness of the problem and make the algorithm solvable. A dataset of 12 images is given with the ground truth of shadings and reflectance, so that the objective evaluations can be conducted. The experiments demonstrate the effectiveness of proposed method.
\end{abstract}

\section{Introduction}

The observed spectrum of a single pixel is determined by illumination, reflectance
and shading. 
Shading image contains illumination condition and geometry information, while reflectance image contains the color information and material reflectance property, which are invariable to light condition and shadow effect. 
The decomposition problem has been a long standing problem of assorted areas
such as both computer graphics and computer vision applications. For instance, shape-from-shading algorithms could benefit from an image with only shading effects, while image segmentation would be easier in a world without cast shadows.

Obviously, intrinsic image decomposition is an ill-posed problem, since there are more unknowns than observations. In order to solve this problem, many work~\cite{shen2008intrinsic, shen2013intrinsic, chen2017intrinsic} focus on sparse representation spatially, but this does not hold for images in general. This paper addressed the problem of the recovery of reflectance and shading from a single multspectral image, namely, the Intrinsic Image Decomposition problem of a whole multispectral image captured under general spectral illumination, hereafter referred to as the IID problem. This problem is worth exploring since geometry and color information are useful under certain circumstances, but one of them always interferes the detection of the other one. Unfortunately, growing dimension of data makes this problem harder to cope with. 

\begin{figure*}[htbp]
\centering                                                          
\subfigure{                    
 \includegraphics[width=0.9\linewidth]{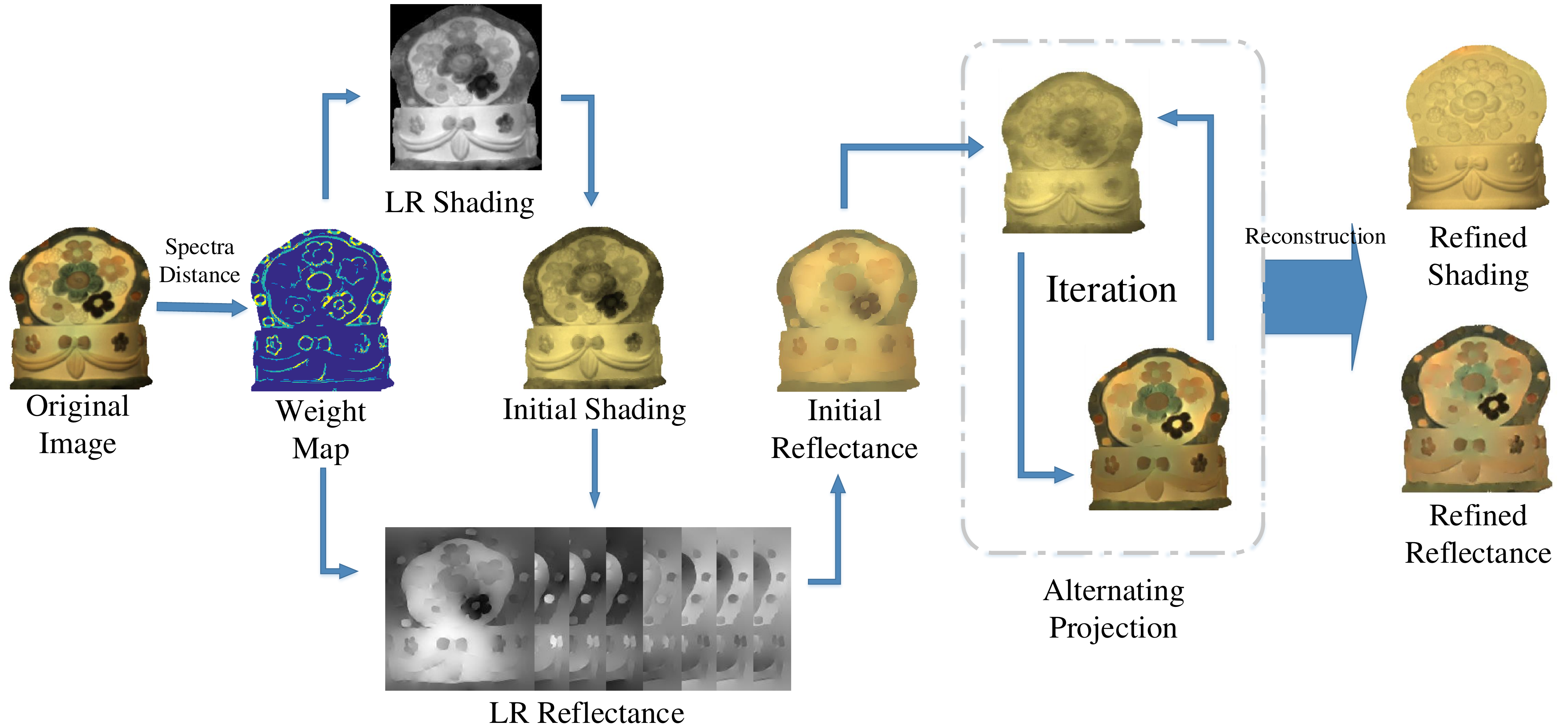} 
}              
\caption{The flowchart of our proposed LRIID algorithm.} 
\label{fig_pipe}
\end{figure*}

The low rank constraint that we proposed is based on the low rank prior, or low rank nature of both shading and reflectance images. According to the inherent nature of the multispectral image, we derive shading basis on the knowledge of illumination condition and derive reflectance subspace bases by means of principle component analysis (PCA) of Munsell color board. Assuming that the basis of Retinex theory would continue to take effect on multispectral domain, we proposed an low-rank-based model so that deriving reflectance and shading from a multispectral image can be modelled as an convex optimization problem. In a significant departure from the conventional approaches which operate in the logarithmic domain, we directly operate on the image domain to avoid adding additional noise or breaking low rank nature. The flowchart of our proposed algorithm for LRIID is shown in Fig.~\ref{fig_pipe}.

Suffering from the lack of ground truth data of shading and reflectance, we provide a ground-truth dataset for multispectral intrinsic images, which provides us a possible way to judge the quality of the decomposition results. Quantitative and qualitative experiments on our dataset have demonstrated that the performance of our work are better than prior work in multispectral domain. Our work can bring merits to multiple applications, such as recolorization, relighting, scene reconstruction and image segmentation.

Our major contribution can be summarized as follows: (1) we extend the Retinex model to multispectral image intrinsic decomposition problem, and propose a low rank constraint to handle the ill-posedness of the problem; (2) we provide a ground-truth dataset for multispectral intrinsic images, which can facilitate future evaluation and comparison of multispectral image intrinsic decomposition algorithms; (3) the proposed method achieves promising results on various of images both quantitatively and qualitatively.

\section{Related Work}

\paragraph{Intrinsic Image Decomposition.} The problem of Intrinsic Image Decomposition (IID) was first introduced by Barrow and Tenenbaum~\cite{barrow1978computer}. The reflectance describes the illumination-invariant albedo of the surface, while the shading contains surface geometric information and illumination condition.

Assorted methods take advantage of additional information, including images sequences~\cite{weiss2001deriving, laffont2013rich} and videos~\cite{lee2012estimation} to avoid shadow effect in poor lighting condition. With the improvement of sensing devices like kinect, depth cue~\cite{barron2013intrinsic, chen2013simple, lee2012estimation} or surface normal~\cite{newcombe2011kinectfusion} have been applied to strengthen their assumption. More recently, Bousseau \etal.~\cite{bousseau2009user} proposed a user-assisted method to further improve the result of separation. 

Lots of methods with single input image are also proposed for the separation task. Bell \etal.~\cite{bell2014intrinsic} developed a dense conditional random field (CRF) based intrinsic image algorithm for images in the wild. Barron \etal.~\cite{barron2015shape} introduced “shape, illumination and reflectance from shading” (SIRFS) model which performs well on images of segmented objects. Sai \etal.~\cite{bi20151} proposed L1 Image Transform model for scene-level intrinsic decomposition. Entropy method~\cite{finlayson2004intrinsic} raised by Finlayson \etal. offered us a new viewpoint to understand this problem. With the abundance and availability of datasets and the development of computational equipment, training-based models~\cite{ bell2001learning,tappen2003recovering,tappen2006estimating,zhou2015learning} have been built to derive reflectance and shading from images.

An especially well-know and wide-employed model called Retinex~\cite{land1971lightness} made an assumption that the large chromatically change is generally caused by changes in reflectance rather than shading. With Retinex theory, we are able to pinpoint where the reflectance changes in local area. Horn, and Funt
and Drew~\cite{ho1990separating} analyzed local derivatives for distinguishing
between image variations that are due to shading or reflectance. But it neglects the connection between pixels sharing the same neighborhood. On the basis of Retinex thoery, we follow work of Chen \etal.~\cite{chen2017intrinsic} to handle intrinsic image decomposition task in multispectral domain.

\paragraph{Sparse Representation.} Researchers have also extended trichromatic color constancy model to multispectral images in order to separate reflectance and shading in higher spectral dimensions. A lot of trials have been made to explore this area. For example, Ikari \etal.~\cite{ikari2008separating} showed us the possibility of separating dozens of multispectral signals. Huynh \etal. ~\cite{huynh2010solution} assumed that the scene could be segmented into several homogeneous surface patches, and were able to estimate the illumination and reflectance spectra under the dichromatic reflectance model. In remote sensing area, Kang \etal.~\cite{kang2015intrinsic} fit multispectral data into trichromatic model to extract features.

To overcome the drawbacks of ambiguity in local analysis, lots of research have been done to reduce the ambiguity of both reflectance and shading. Shen \etal.~\cite{shen2008intrinsic} proposed a global optimization algorithm which combines Retinex theory and non-texture constraint to obtain global consistency of image structures. Shen~\cite{shen2013intrinsic} further applied sparse representation of reflectance as global constraint of their observation. Material cues~\cite{nadian2016intrinsic} has also been introduced. As to multispectral images, Chen \etal.~\cite{chen2017intrinsic} used super-pixel to cut down the number of unknown parameters in this underdetermined problem. Unlike the approaches above, we assumed that both shading and reflectance live in low dimensional subspace. the low rank nature of shading space is widely acknowledged and exploited in prior work~\cite{ho1990separating}, and the low-dimensional subspace model of reflectance is introduced by~\cite{maloney1986evaluation, parkkinen1989characteristic, zheng2015illumination}. With the help of training data of reflectance bases and illumination spectra, we can solve this problem effectively.

\paragraph{Dataset.} As far as establishing ground-truth for intrinsic images,
Tappen \etal.~\cite{tappen2003recovering} created small sets of both computer-generated
and real intrinsic images. The computer-generated images
consisted of shaded ellipsoids with piecewise-constant reflectance. The real images were created using green marker on crumpled paper~\cite{tappen2006estimating}. Grosse \etal.~\cite{grosse2009ground} provided an all-rounded dataset which is widely used in following analysis. Bell \etal.~\cite{bell2014intrinsic} also introduced Intrinsic Images in the Wild, a large scale, public dataset for evaluating intrinsic image decompositions of indoor scenes. In multispectral field, \cite{yasuma2010generalized, chakrabarti2011statistics} provided a set of multispectral images of various objects without ground truth, and Chen \etal.~\cite{chen2017intrinsic} built a dataset but the number, spectral resolution and diversity of images are limited.
There is no other attempts to establish ground truth for multispectral intrinsic images.

\section{Our Model}
We assume the object surface as Lambertian and hence has diffuse reflection. 
In most prior work on intrinsic image decomposition, the captured luminance 
spectrum at every point $l_p$ is modelled as the product of Lambertian reflectance spectrum $r_p$ and shading spectrum $s_p$, where $s_p$ is used to characterize the combined effect of object geometry, illumination, occlusion and shadowing. Mathematically, this model can be expressed as
\begin{equation}
l_p = s_p.*r_p
\end{equation}
where $l_p$, $r_p$ and $s_p$ are all vectors with dimensions equal to the number of spectral bands of the captured image, “$.*$” denotes element-wise multiplication. The problem is to derive $s_p$ and $r_p$ from observed multispectral luminance vector $l_p$: In this project, we will focus on recovering the reflectance spectrum using this model. Once $r_p$ is determined, the shading image can be derived by point-wise division.

Different from the conventional approaches which operate in the logarithmic domain, we directly formulate the problem in the image domain, and this can overcome numerical problems caused by the logarithmic transformation of the image values, where noise in pixels with low intensity values can lead to large variations. Besides, although there have been substantial evidence of the low rank nature of the reflectance space, it is not clear whether the logarithmically transformed reflectance space is still low rank. This makes it hard to incorporate the low rank prior in formulations based on log-transformed images. 

\subsection{Estimate Reflectance or Shading Independently}

The Retinex model makes following two important observations:
\begin{enumerate}[1)]
\item When there is significant reflectance change between two adjacent pixels $p$ and $q$, the shading is typically constant. This leads to the relation $l_p./l_q$ = $r_p./r_q$, where “$./$” denotes element-wise division; 
\item When the expected reflectance difference between two pixels is small, the recovered reflectance difference between the two pixels should be small.
\end{enumerate}

First we look at whether we can separate reflectance and shading from the measured luminance signal independently. Take reflectance for example. By recognizing two adjacent pixels which have the same shading, the ratio relationship can be written as $l_p.*r_q = l_q.*r_p$, or $L_pr_q = L_qr_p$ where $L_p$ is a diagonal matrix consisting of spectral elements in $l_p$, we formulate the energy functions in terms of the reflectance vectors directly:
\begin{equation}
\begin{array}{c}
{Esc = \sum\limits_{p, q \in \mathcal{N}_{sc}}\lVert w_{p,q}(L_pr_q-L_qr_p) \rVert^d} \\
\\
{Erc = \lVert v_{p,q}(r_p-r_q) \rVert^d}
\end{array}
\label{EscErc}
\end{equation}

where $\mathcal{N}_{sc}$ and $\mathcal{N}_{rc}$ denote neighborhood pair sets and $w_{p, q}$ and $v_{p, q}$ denote weights. $w_{p, q}$ would be large but $v_{p, q}$ will be small when the expected reflectance difference between two adjacent pixels $p$ and $q$ are large, and vice verse. To make the formulation general, we use d to indicate the error norm, with $d = 2$ for L2 norm, and $d = 1$ for L1 norm. L1 norm is more difficult to solve, but the
solution can be more robust to outliers.

If we directly solve for $r_p$, the above energy function can be written as the sum of $K$ terms, one for each
spectral component and each term can be separately minimized. With a little exercise, it can be shown that
the minimal is achieved exactly when $r_p = l_p$; This is due to the inherent ambiguity of the problem, when no other constraints are imposed on $r_p$. we reduce the ambiguity by exploiting the fact that the reflectance spectra of typical object surfaces live
in a low dimensional subspace of $R^K$, so that any reflectance vector can be written as a linear combination
of $J_r$ basis, with $J_r < K$.

Let $B_r$ represent the $K \times J$ basis matrix for representing the reflectance vector, $r_p$ can be written as $r_p = B_r*\widetilde{r}_p$. The energy function in Eq.(\ref{EscErc}) now becomes:
\begin{equation}
\begin{array}{c}
{Esc = \sum\limits_{p, q \in \mathcal{N}_{sc}}\lVert w_{p,q}(L_pB_r\widetilde{r}_q-L_qB_r\widetilde{r}_p) \rVert^d} \\
\\
{Erc = \lVert v_{p,q}(B_r\widetilde{r}_p-B_r\widetilde{r}_q) \rVert^d}
\end{array}
\label{sparseEscErc}
\end{equation}

The combined energy can be represented in a matrix form as:
\begin{equation}
E = \lVert W_{L, B_r}\widetilde{R} \rVert^d + \lambda_1 \lVert V_{B_r}\widetilde{R} \rVert^d
\end{equation}

where $\widetilde{R}$ consists of $\widetilde{r}_p$ for all pixels in a vector. The matrix $W_{L, B_r}$ depends on the neighborhood $\mathcal{N}_{sc}$
considered, the weight $w_{p, q}$, the reflectance basis $B_r$ used, and importantly the luminance data $l_p$; whereas the matrix $V_{B_r}$ depends on the neighborhood $\mathcal{N}_{rc}$ considered, the weight $v_{p,q}$ and the reflectance basis
$B_r$ used. Therefore, with this non-logarithmic formulation, we encode the constraint due to the measured luminance data in the matrix $W_{L, B_r}$.

The ambiguity with the scaling factor is inherent in all intrinsic image decomposition problems since only the product of reflectance and shading is known. To circumvent the ambiguity about the scaling factor, we also explored another solution, where we express the generic constraint on the coefficient sum as $M\widetilde{R} = C$, and augment the original energy function to enforce this constraint:
\begin{equation}
\begin{aligned}
E_{\text{refl}} = \lVert W_{L, B_r}\widetilde{R} \rVert^d + \lambda_1 \lVert V_{B_r}\widetilde{R} \rVert^d + \lambda_2 \lVert M_r\widetilde{R}-C \rVert^d
\end{aligned}
\label{E_ref}
\end{equation}

Similarly, the low rank nature of the shading space is also widely acknowledged and exploited in prior work [14]. Shading is inherently low rank, because there are usually only a few lighting sources with different illumination spectra acting in each captured scene, and the shading effect due to geometry and shadowing only modifies the spectra by a location-dependent scalar. If there is a single illumination source and its spectrum is known or is able to be identified by method of~\cite{zheng2015illumination}, we will use this spectrum(after normalization) as the only shading basis vector($J_s = 1$ and $B_s$ equals to this normalized spectrum). Likewise the problem can be formulate as minimize
\begin{equation}
\small
\begin{aligned}
E_{\text{shad}} 
& = \lVert W_{B_s}\widetilde{S} \rVert^d + \lambda_1 \lVert V_{L, B_s}\widetilde{S} \rVert^d + \lambda_2 \lVert M_s\widetilde{S}-C \rVert^d
\end{aligned}
\label{E_shad}
\end{equation}

\subsection{Simultaneous Recovery of Reflectance and Shading}
Based on the formulation that solves reflectance and shading respectively, we proposed an optimization algorithm that simultaneously solves both shading and reflectance. We assume that the low rank subspace of the shading and reflectance are known, represented
by basis matrices $B_s$ and $B_r$, respectively, so that $s_p = B_s\widetilde{s}_p$ and $r_p = B_r\widetilde{r}_p$. We will use $\widetilde{S}$ to denote the
long vector consisting of shading coefficient vectors $\widetilde{s}_p$ at all pixels, and $\widetilde{R}$ the long vector consisting of
reflectance coefficient vectors $\widetilde{r}_p$. We propose to solve $\widetilde{s}_p$ and $\widetilde{r}_p$, or equivalently $\widetilde{S}$ and $\widetilde{R}$, by minimizing a weighted average of the following energy terms. 

When shading is expected to be similar in pixels $p$ and $q$, we have $s_p \approx s_q $ and $l_p.*r_q \approx l_q.*r_p$, or $L_pr_q = L_qr_p$, where $L_p$ is a diagonal matrix consisting of spectral elements in $l_p$. We formulate the energy functions directly:
\begin{equation}
\begin{aligned}
E_{sc} &= \sum\limits_{p, q \in \mathcal{N}_{sc}}\lVert w_{p,q}(L_pr_q-L_qr_p) \rVert^d + \lVert w_{p,q}(s_p-s_q) \rVert^d \\
&=\lVert W_{L, B_r} \widetilde{R} \rVert^d + \lVert W_{B_s} \widetilde{S} \rVert^d
\end{aligned}
\end{equation}

When reflectance is expected to be similar in pixels $p$ and $q$, we have $r_p \approx r_q $ and $l_p.*s_q \approx l_q.*s_p$, leading to a regularization energy
\begin{equation}
\begin{aligned}
E_{rc} &= \sum_{p, q \in \mathcal{N}_{rc}}\lVert v_{p,q}(L_ps_q-L_qs_p) \rVert^d + \lVert v_{p,q}(r_p-r_q) \rVert^d\\
&= \lVert V_{L, B_s} \widetilde{S} \rVert^d + \lVert V_{B_r} \widetilde{R} \rVert^d
\end{aligned}
\end{equation}

The inherent data constraint $l_p = s_p.*r_p$ leads to another energy function:
\begin{equation}
\begin{aligned}
E_{\text{data}}&= \sum_{p}\lVert s_p.*r_p- l_p\rVert^d = \lVert Q_{\widetilde{S}, B_s, B_r} \widetilde{R} - L \rVert^d \\
&= \lVert Q_{\widetilde{R}, B_r, B_s} \widetilde{S} - L \rVert^d
\end{aligned}
\label{data_cons}
\end{equation}

where $Q_{\widetilde{S}, B_s, B_r}$ is a block diagonal matrix that depends on the solution for $\widetilde{R}$ and the basis matrices $B_s$ and $B_r$(likewise $Q_{\widetilde{R}, B_r, B_s}$).

The problem is to find $\widetilde{S}$ and $\widetilde{R}$
that minimizes a weighted average of the three energy functions:
\begin{equation}
\small
\begin{aligned}
E &= E_{sc}+\lambda_{1}E_{rc} + 2\lambda_{data} E_{data} \\
& = \lVert W_{L, B_r} \widetilde{R} \rVert^d + \lVert W_{B_s} \widetilde{S} \rVert^d +\lambda_{1} \left(\lVert V_{L, B_s} \widetilde{S} \rVert^d + \lVert V_{B_r} \widetilde{R}\rVert^d \right) \\
&+ \lambda_{data} \lVert Q_{\widetilde{S}, B_s, B_r} \widetilde{R} - L \rVert^d + \lambda_{data} \lVert Q_{\widetilde{R}, B_r, B_s} \widetilde{S} - L \rVert^d\\
\end{aligned}
\label{Etotal}
\end{equation}

Direct solution of the above problem solving $s_p$ and $r_p$ simultaneously is hard because of the bilinear nature of the data term. We applied iterative solution, where we solve $\widetilde{R}$ and $\widetilde{S}$ iteratively using alternating projection. As the dimension of the shading subspace is likely to be smaller than the dimension of the reflectance subspace, we solve the shading $\widetilde{S}$ first. Also there
are typically more subregions in an image with similar reflectance, where it is easier to use the constant reflectance constraint to resolve the ambiguity about shading.

The only problem is that we need to give a initial estimation of $\widetilde{S}$ and $\widetilde{R}$ in order to effectuate the data constraint in Eq.(\ref{data_cons}). With the generic constraint in Part 3.1, Eq.(\ref{E_ref}) and Eq.(\ref{E_shad}), we give a previous estimation of $\widetilde{S}$ first, and then we can update $\widetilde{R}$ and find reflectance basis matrix, finally $\widetilde{R}$ and $\widetilde{S}$ are estimated respectively using the data constraint in Eq.(\ref{data_cons}).

More specifically, the whole recovery algorithm can be classified as Algorithm~\ref{algo}:
\begin{algorithm}[]
\label{algo}
\caption{LRIID algorithm}
\setcounter{AlgoLine}{0}
\textbf{Step 1}: Assign constant-shading weights $w_{p,q}$ and constant-reflectance weights $v_{p,q}$.\\
\textbf{Step 2}: Solve an initial low rank estimate of the shading image $\widetilde{S}$ using a generic constraint.\\
\textbf{Step 3}: Solve an initial low rank reflectance estimate $\widetilde{R}$ using a generic constraint and the data constraint defined by the previous shading estimate.\\
\Repeat{until the solution for $\widetilde{S}$ and $\widetilde{R}$ converge}{
\textbf{Step 4}: Solve $\widetilde{S}$ using the data constraint defined by the previous reflectance estimate.\\
\textbf{Step 5}: Solve $\widetilde{R}$ using the data constraint defined by the previous shading estimate.
}
\textbf{Step 6}: Reconstruct $S$ and $R$ to get the refined shading and reflectance.
\end{algorithm}

\section{Details}

\subsection{Weight Choice}

As we know, images suffering from poor light condition may contain shadow area, which would in turn bring in unnecessary edges that confuse the algorithm. Various methods are used to determined weights $w_{p, q}$ and $v_{p, q}$, including pixel gradient~\cite{funt1992recovering, kimmel2003variational, land1971lightness}, hue~\cite{zhao2012closed}, correlation between vectors~\cite{jiang2010correlation} and learning~\cite{tappen2003recovering}. we proposed a illumination-robust and compute-friendly distance -- normalized cosine distance, to signify the differences between spectra of pixels in one neighborhood. This distance can be formulated as

\begin{equation}
d_{p,q \in \mathcal{N}_{sc}} = 1-\frac{l_p' l_q}{\lvert l_p \rvert \cdot \lvert l_q \rvert}
\end{equation}

$d_{p, q}$ tends to approximate to 0 when pixel $p$ and $q$ have same spectra, and depart from 0 when spectra of $p$ and $q$ are different. In order to derive weight $w_{p, q}$, we need to further magnify the difference between homogeneous and heterogeneous pixels and make it more robust to the noise. In our implementation

\begin{equation}
\begin{array}{ccc}
{w_{p, q} = \frac{1}{1+e^{\alpha(d_{p, q}-\beta)}}}   & &
{v_{p, q} = 1-w_{p, q}}
\end{array}
\end{equation}

$\alpha$ and $\beta$ are parameters of sigmoid function. To set $\alpha$ and
$\beta$, We sample 20 value of $\alpha$ within [1000, 10000] and 50 values of $\beta$ within [$10^{-5}$, $10^{-2}$] and choose which perform best.

\begin{figure*}[htbp]
\centering                                                          \subfigure[LMSE = 0.050, $\alpha = 1000, \ \beta = 0.01$]{                    
\begin{minipage}{5.5cm}
\centering                                                          \includegraphics[width=1\linewidth]{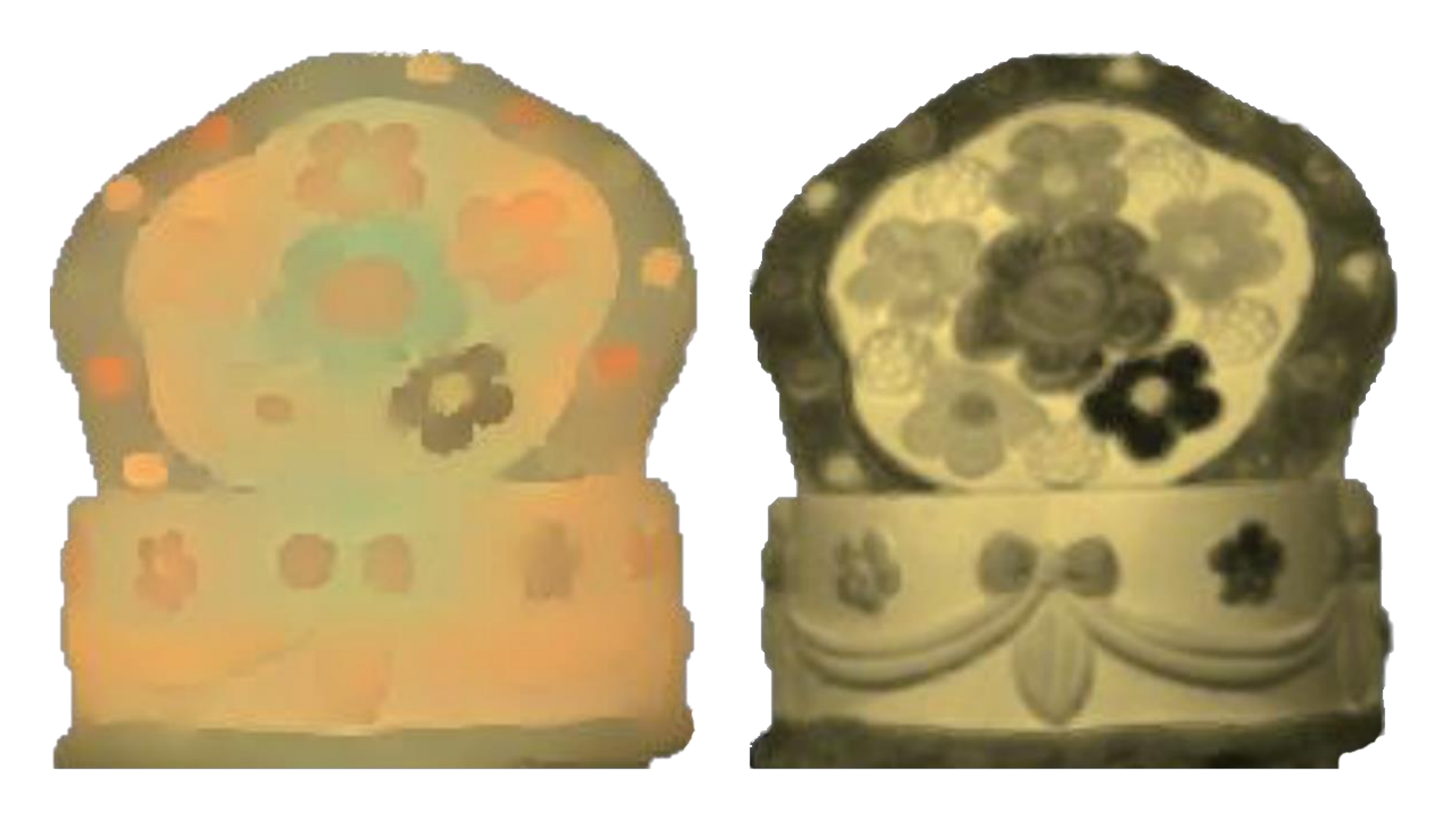}               
\end{minipage}}
\subfigure[LMSE = 0.026, $\alpha = 5000, \ \beta = 0.0032$]{                    
\begin{minipage}{5.5cm}
\centering                                                          \includegraphics[width=1\linewidth]{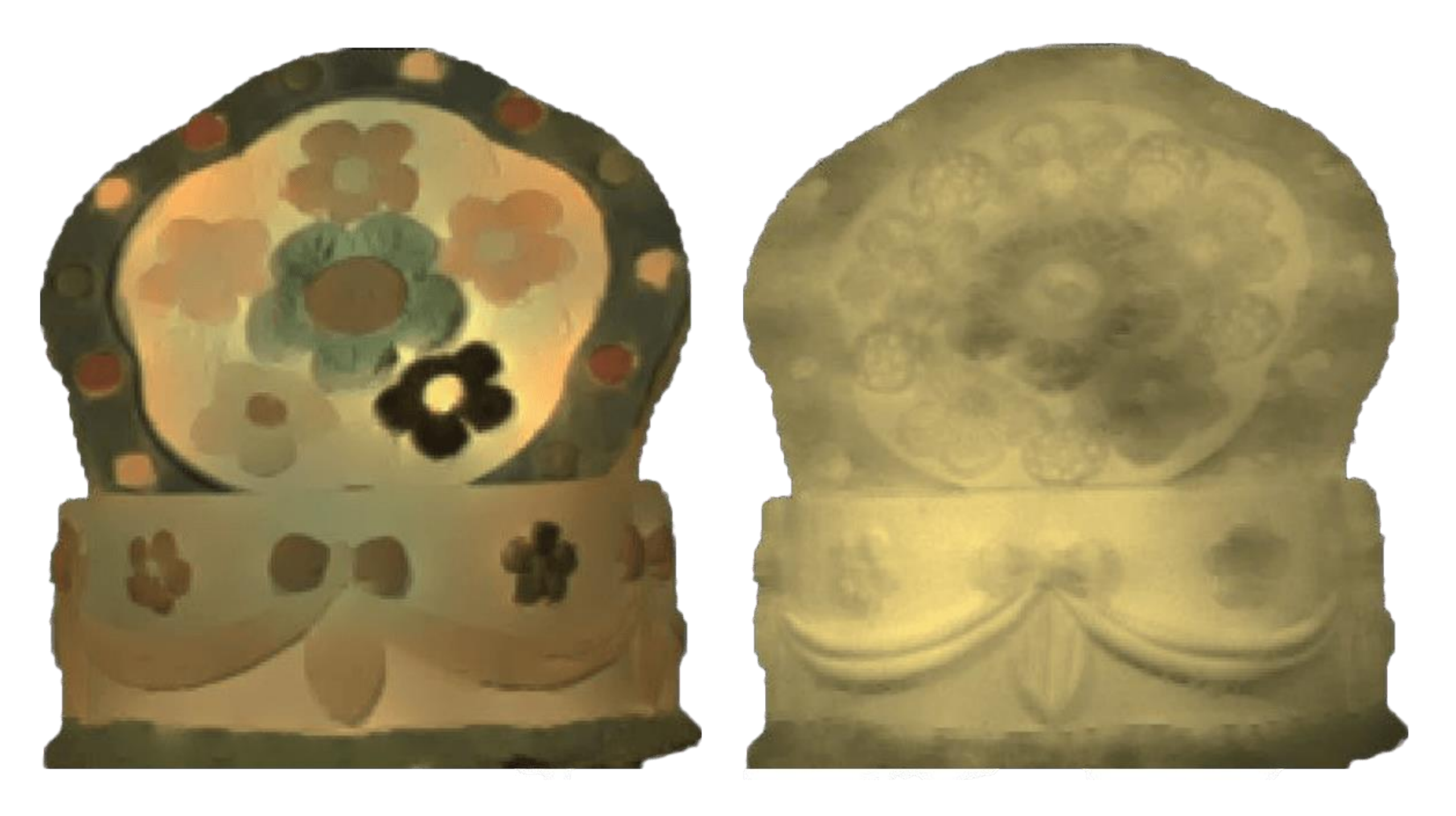}                
\end{minipage}}
\subfigure[LMSE = 0.031, $\alpha = 8000, \ \beta = 7.9e-4$]{                    
\begin{minipage}{5.5cm}
\centering                                                          \includegraphics[width=1\linewidth]{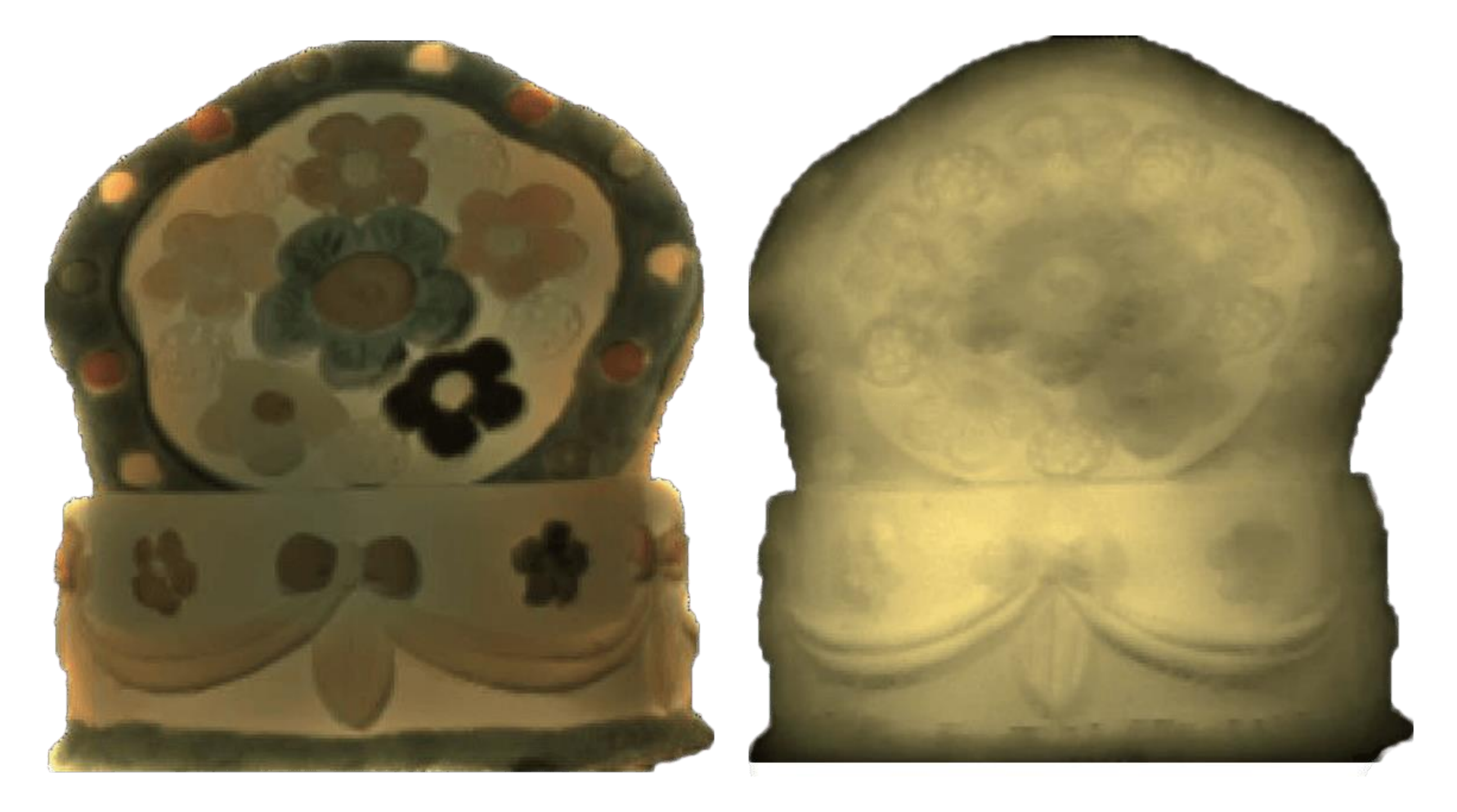}                
\end{minipage}}
\caption{Results of threshold $\alpha$ and $\beta$. (b) achieves good result while shading and reflectance overlap clearly in (a) and (c).} 
\label{fig_weight}
\end{figure*}

Fig.~\ref{fig_weight} shows different separation results when $\beta$ changes. Local mean squared error (LMSE) is valued in 3 results. If $\beta$ is too small, shading tends to be more blurred; while $\beta$ is too big, reflectance would be blurred.

\subsection{Low Rank Constraint}

An important step to formulate a low rank constraint is to derive the low rank basis. With the help of multispectral imaging systems as PMIS~\cite{cao2011prism} and CASSI~\cite{arce2014compressive}, we can successfully get grouth-truth illumination spectra. Also, there are plenty of work referring to how to extract illumination from the image. For example, ~\cite{zheng2015illumination} can be applied in multispectral domain and performs well in implementation. In order not to complicate our method, We simplify the process of getting normalized illumination spectra $B_s$ by using the grouth-truth illumination data.

When it comes to the low rank basis of reflectance, the authors of~\cite{maloney1986evaluation,parkkinen1989characteristic} have found $J_r$ to be around 8 so as to reach the best trade-off between expression power and noise resistance in the process of fitting reflectance spectra.we set $J_r$ to be 8 and use the matrix introduced by~\cite{maloney1986evaluation} and perform Principle Component Analysis (PCA) to derive $B_r$ from it.

\subsection{Initial Estimation}

In our implementation, only horizonally adjacent and vertically adjacent pixels are considered in the neighborhood $N_{sc}$ and $N_{rc}$. Suppose there are $N$ pixels in the image. The size of matrices $W_{L, B_r}$, $V_{B_r}$ and $\widetilde{R}$ are $4NK \times NJ_r$, $4NK \times NJ_r$ and $NJ_r \times 1$ respectively in Eq.(\ref{sparseEscErc}). Similarly, The size of matrices $W_{B_s}$, $V_{L, B_s}$ and $\widetilde{S}$ in Eq.(\ref{E_shad}) are $4NK \times NJ_s$, $4NK \times NJ_s$ and $NJ_s \times 1$ respectively.

To avoid ambiguity, we further require that the shading image has small deviation from the input image. In Eq.(\ref{E_ref}), let $M_r$ be a identity matrix, and $C$ be a long vector concatenating all the pixels of the original image.

We use L2 form for all terms. so that the above problem for solving Eq.(\ref{E_ref})or Eq.(\ref{E_shad}) is a quadratic programming problem, and can be solved efficiently using conjugated gradient method. In Algorithm \ref{algo}, The solution to the unconstrained optimization problem in Step 2 satisfies the following linear equation
\begin{equation}
\footnotesize
\begin{aligned}
\lambda_2 M_s^T C = \left( W_{B_s}^T W_{B_s} + \lambda_1 V_{L, B_s}^T V_{L, B_s} + \lambda_2 M_s^T M_s \right)\widetilde{S} = Q_s\widetilde{S}
\end{aligned}
\end{equation}

In Step 3, a data constraint defined by shading estimate need to be added to Eq.(\ref{E_ref}), so that the linear equation can be written as
\begin{equation}
\footnotesize
\begin{aligned}
&\lambda_{\text{data}} Q_{\widetilde{S}}^T L + \lambda_2 M_r^T C = \\
& \left( W_{L, B_r}^T W_{L, B_r} + \lambda_1 V_{B_r}^T V_{B_r} + \lambda_2 M_r^T M_r +\lambda_{\text{data}} Q_{\widetilde{S}}^T Q_{\widetilde{S}}\right)\widetilde{R} = Q_r\widetilde{R}
\end{aligned}
\end{equation}

Because the matrix $Q_r$ and $Q_s$ is self-adjoint and sparse, we can solve this equation iteratively, which typically converges very fast. Here, $\lambda_1$ and $\lambda_2$ are positive weights for combining three different objective functions. In our implementation, we set $\lambda_1 = 2$, $\lambda_2 = 0.01$ and $\lambda_{\text{data}} = 1$ empirically.

\subsection{Iteration Performance}

We use alternating projection to get refined shading and reflectance. Just like what we stated in Step 4 and Step 5 in Algorithm \ref{algo}, we we update shading first and then the reflectance. In each round of iteration, gradient descent method is applied. Eq.~\ref{Etotal} converges whenever it reaches the maximum iteration times 1000 or $\nabla E < 0.01$.

\begin{figure}[htbp]
   \includegraphics[width=1\linewidth]
                   {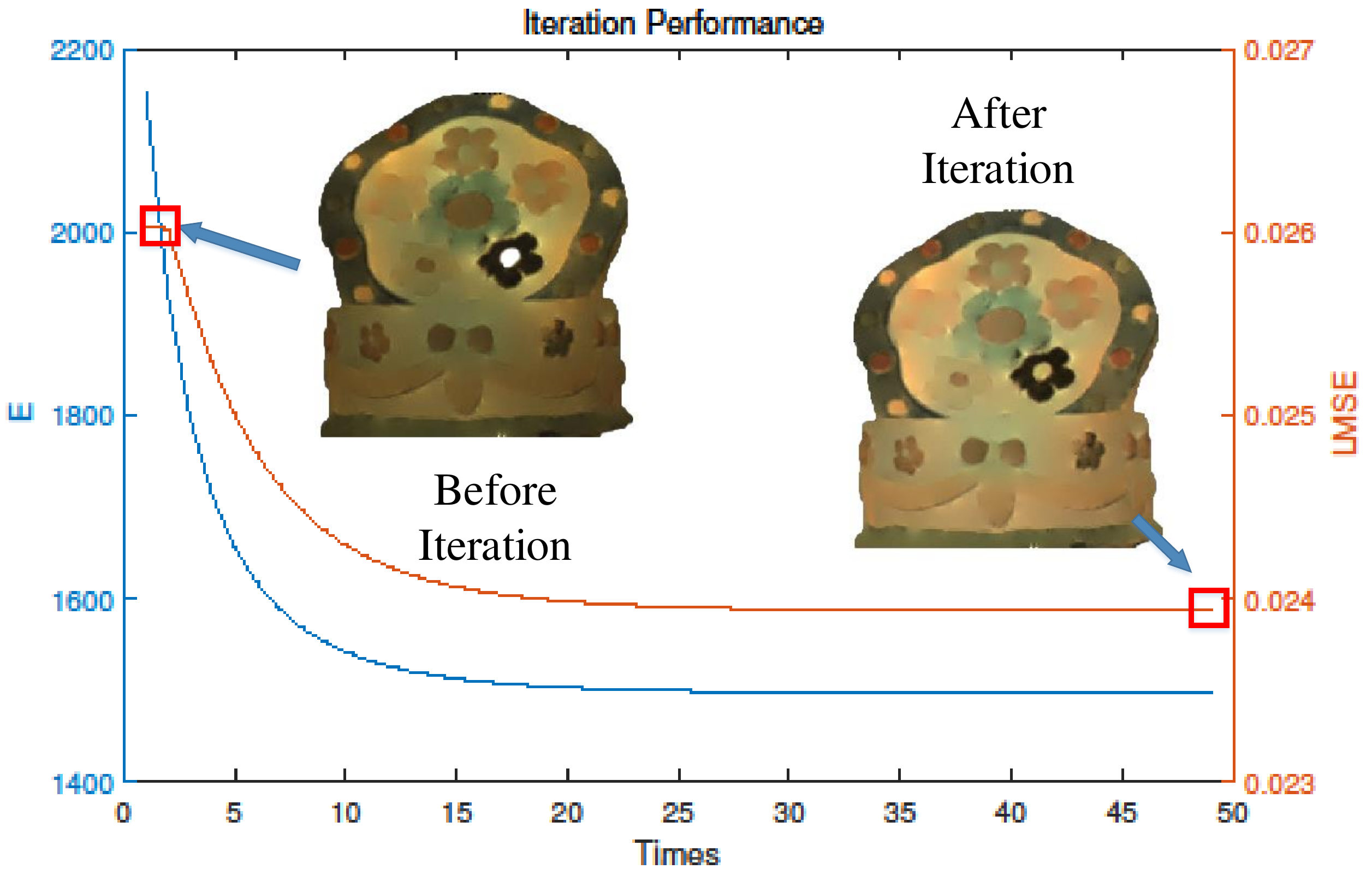}
\caption{Iteration Performance. The algorithm converges within 50 times. Two images demonstrate the reflectance image which generated before and after iteration.}
\label{iter_curve}
\end{figure}

Fig.~\ref{iter_curve} demonstrates the iteration performance of our algorithm. Both cost function and LMSE decrease via iteration. Before the iteration, some shadow which should be the component of the shading image remains on the reflectance, while the brightness of the reflectance image tends to be uniform after iteration.

\section{Experimental Results}

In this section, we provide extensive experimental validation of the proposed method. For the better visualization, we show the result in pseudo-rgb and linearly normalized the image to the range [0, 1]. We first show the performance of our algorithm. This is followed by our method on on-line dataset and visual results. Finally, we test on our dataset with ground truth and compare our method with~\cite{chen2017intrinsic}.

\subsection{Experiments on proposed dataset}
We provide a benchmark dataset with ground truth for the performance evaluation of multispectral image intrinsic decomposition problem. Following~\cite{grosse2009ground}, we use local mean squared error (LMSE) from the ground truth to measure shading and reflectance image. We also compare with~\cite{chen2017intrinsic}, which proposed intrinsic image decomposition algorithm in multispectral domain.

A benchmark dataset with ground-truth illumination, shading, reflectance and specularity was presented in~\cite{chen2017intrinsic} for performance evaluation of multispectral image intrinsic decomposition. Inspired by their ideas, we build up the newest multispectral intrinsic ground-truth dataset including 12 scenes under the same environmental conditions. We apply the updated mobile multispectral imaging camera to acquire the multispectral scenes, which could provide higher resolutions in spectral data ranging from 450nm-700nm with 118 spectral channels.  Compared with the dataset provided by~\cite{chen2017intrinsic}, ours is a similar but flowering in the diversity of projects with more detailed and bumpy scenarios which enables the dataset show further and potential applications in other vision researches (e.g. IRSS, segmentation and recognition). 

Here, we evaluate our algorithm via our proposed dataset and use LMSE from the ground truth to validate our algorithm quantitatively. Compared with ground-truth, decomposition results that we achieved are desirable in terms of both the LMSE score (0.018 in average for the entire data set and the visual quality of the decomposed reflectance and shading results).

\begin{figure}[htbp]
\centering                                                          
\subfigure{                    
\begin{minipage}{8cm}
\centering                                                          
\includegraphics[width=1\linewidth]{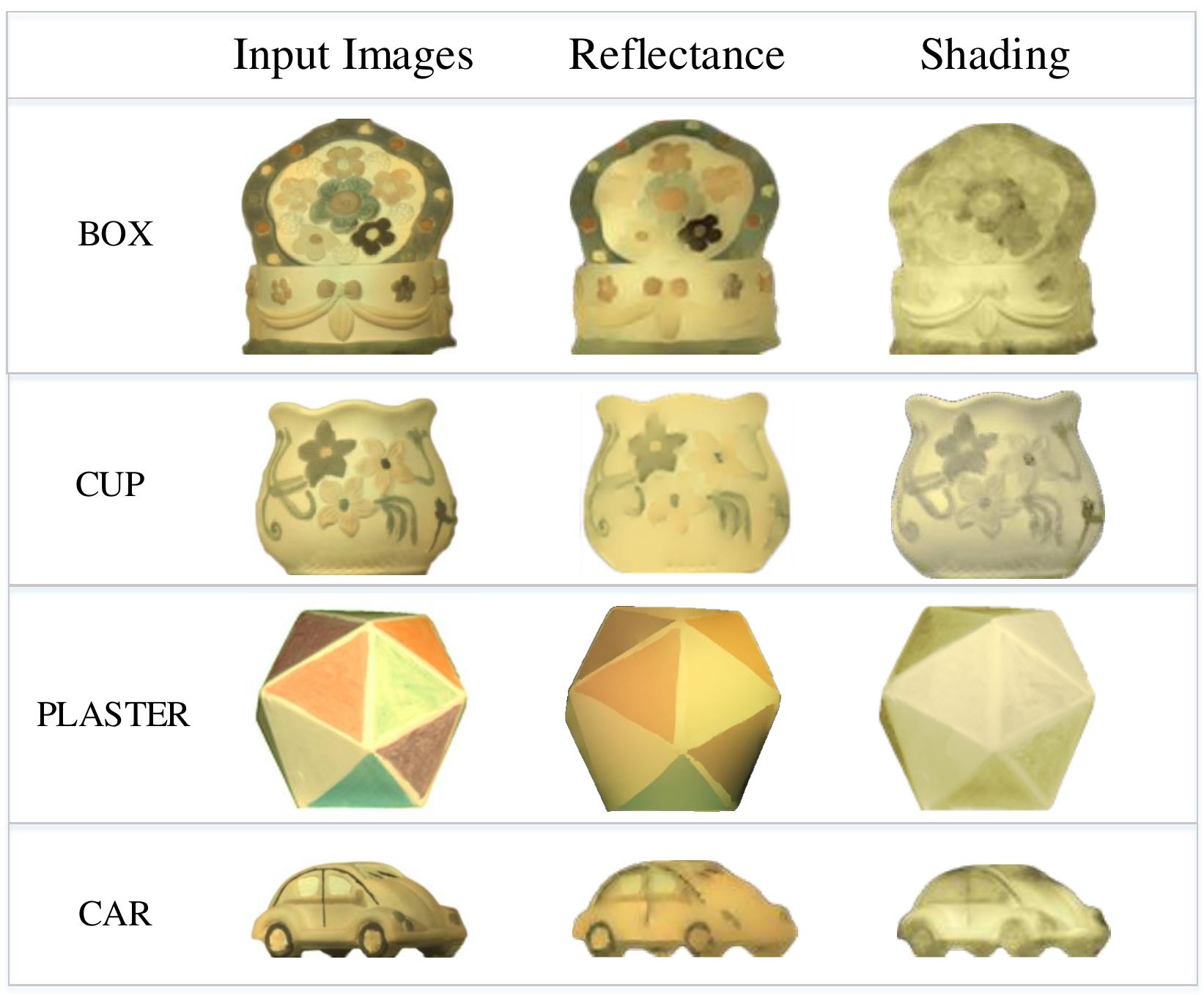}               
\end{minipage}}
\caption{Results on sample images from the benchmark data set. we only show reflectance and shading color images synthesized from spectral data of 4 examples.} 
\label{fig_by_our}
\end{figure}

We display 4 examples from our dataset. For all the input diffuse images equipped with multispectral data without the effect from illumination, the corresponding visualized RGB images for reflectance and shading are listed in Fig.~\ref{fig_by_our} together with the LMSE results. It is clear that our method could produce pleasing visually decompositions based on multispectral images.  In~\cite{chen2017intrinsic} generic constraint forces the sum value of a group of pixels to be a certain constant C, while in our method we let each pixel in shading image approximates to the pixel value of original image in the same position. Our generic constraint matrix is more sparse than that in~\cite{chen2017intrinsic}, so that our algorithm will converge more quickly. What we have to emphasize here is that, we did all the image processing and metric computations of LMSE on down-sampled 30 out of 118 spectral channels (because~\cite{chen2017intrinsic} failed to process more bands on our 32G bytes memory) but all multispectral images are finally visualized above using corresponding synthesized RGB data.

\begin{table}[htbp]
    \caption{Performance statistics for dataset image}

  \centering

    \begin{tabular}{c|c|c|c|c}
    \hline
    \hline
          & \multicolumn{2}{c|}{Time(s)} & \multicolumn{2}{c}{LMSE} \\
\hline          & SIID  & LRIID & SIID  & LRIID \\
    \hline
    box   & 9.221 & 1.351 & 0.032  & \textbf{0.023}  \\
    cup   & 7.610 & 0.675 & 0.016  & \textbf{0.012}  \\
    car1  & 14.771 & 0.876 & 0.025  & \textbf{0.012}  \\
    bottle1 & 11.275 & 0.666 & 0.062  & \textbf{0.031}  \\
    bottle2 & 8.342 & 2.709 & \textbf{0.005}  & 0.008  \\
    bottle3 & 7.993 & 2.337 & \textbf{0.009} & \textbf{0.009}  \\
    bus   & 13.667 & 2.669 & \textbf{0.030}  & 0.031  \\
    car2  & 7.643 & 1.807 & 0.030  & \textbf{0.024}  \\
    dinosaur & 13.465 & 2.306 & \textbf{0.021}  & 0.023  \\
    minion & 17.794 & 2.517 & 0.020  & \textbf{0.018}  \\
    plane & 10.763 & 2.125 & 0.024  & \textbf{0.015}  \\
    train & 9.108 & 0.832 & 0.017  & \textbf{0.015}  \\
    \hline
    Avg.  & 10.971 & \textbf{1.739} & 0.024  & \textbf{0.018}  \\
    \hline
    \hline
    \end{tabular}%
  \label{tab:1}%
\end{table}%

We display the computation time for examples in Fig.~\ref{fig_by_our}. Here SIID denotes spectral intrinsic image decomposition from the latest method based on Retinex in~\cite{chen2017intrinsic}. LRIID denotes the low rank multispectral image intrinsic decomposition from our algorithm. We compare the computation results obtained with and without low rank constraint. It can be seen that the low rank constraint helps to improve the computation time and decomposition results at the same time. 

In table \ref{tab:1}, we demonstrate the performance statics for dataset images. The last row shows the average metrics for the ground truth dataset. On average, the SIID method requires almost 11 seconds to process a multispectral image and generates results with 0.024 LMSE. While our method LRIID takes 1.739 seconds and produce results with 0.018 LMSE. Although we reduce the average LMSE slightly from 0.024 to 0.018, but the computation time is qualitatively more efficient than that of SIID, and the running time approximates to the time in RGB cases in~\cite{shen2008intrinsic}. Moreover, We note that our method is memory-friendly and is able to process larger images with more spectral bands.

\begin{figure}[htbp]

\centering                                                          
\subfigure[GT]{                    
\begin{minipage}{2.1cm}
\centering                                                          
\includegraphics[width=1\linewidth]{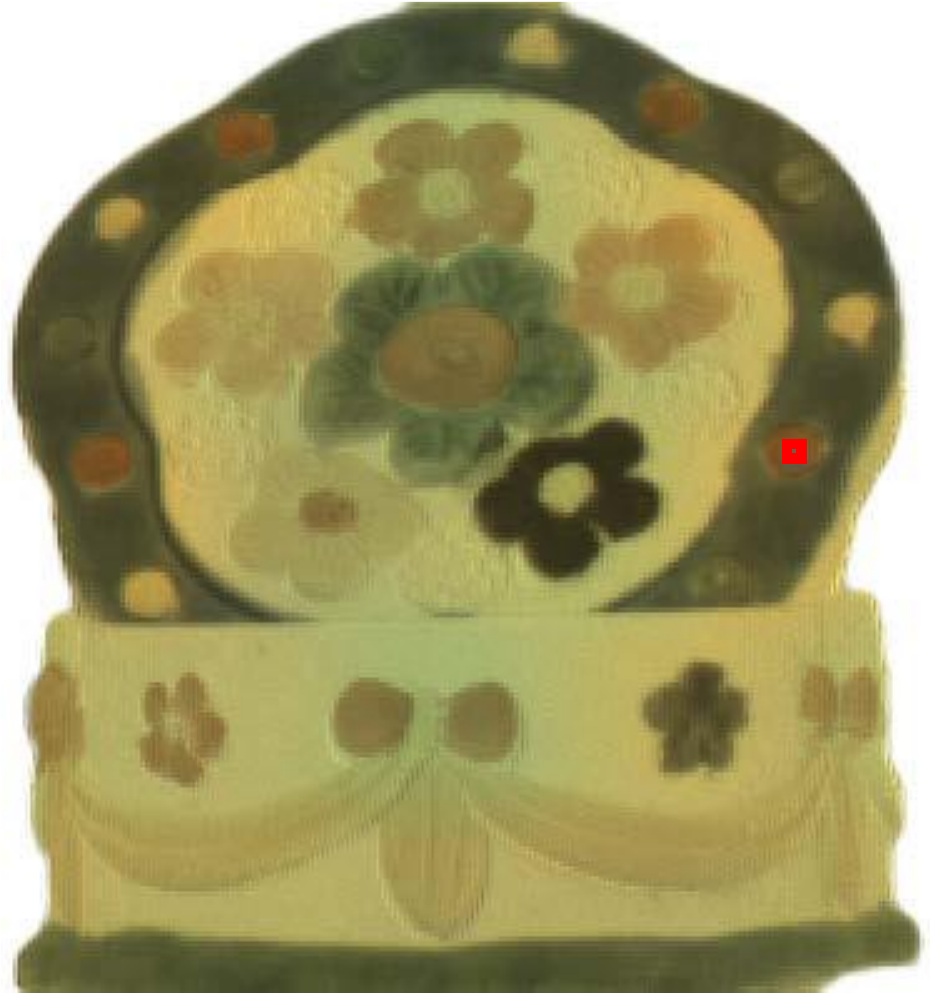}               
\end{minipage}}
\subfigure[Ours]{                    
\begin{minipage}{2.3cm}
\centering                                                          
\includegraphics[width=1\linewidth]{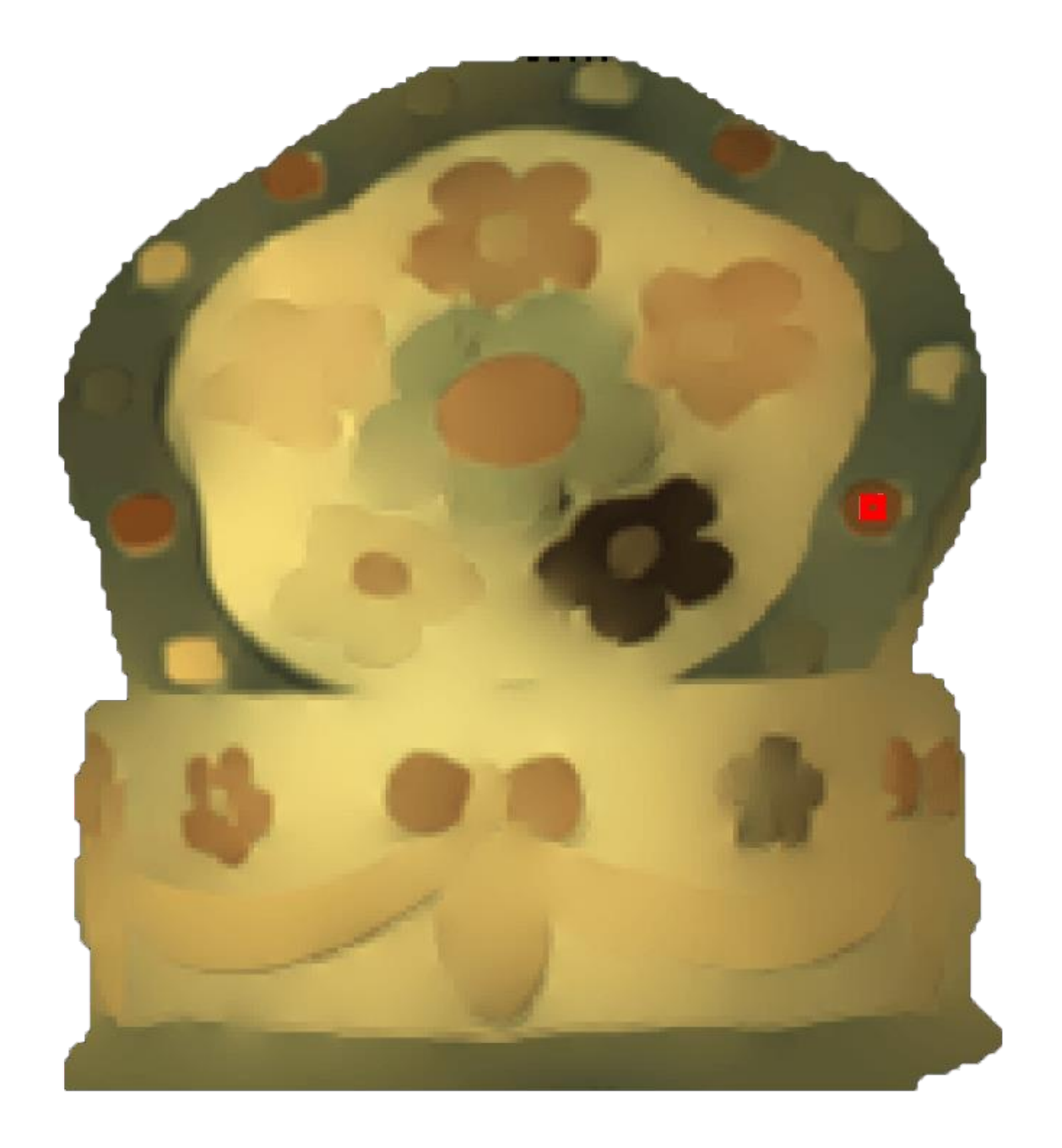}                
\end{minipage}}
\subfigure[Spectra Curve]{                    
\begin{minipage}{2.9cm}
\centering                                                          
\includegraphics[width=1\linewidth]{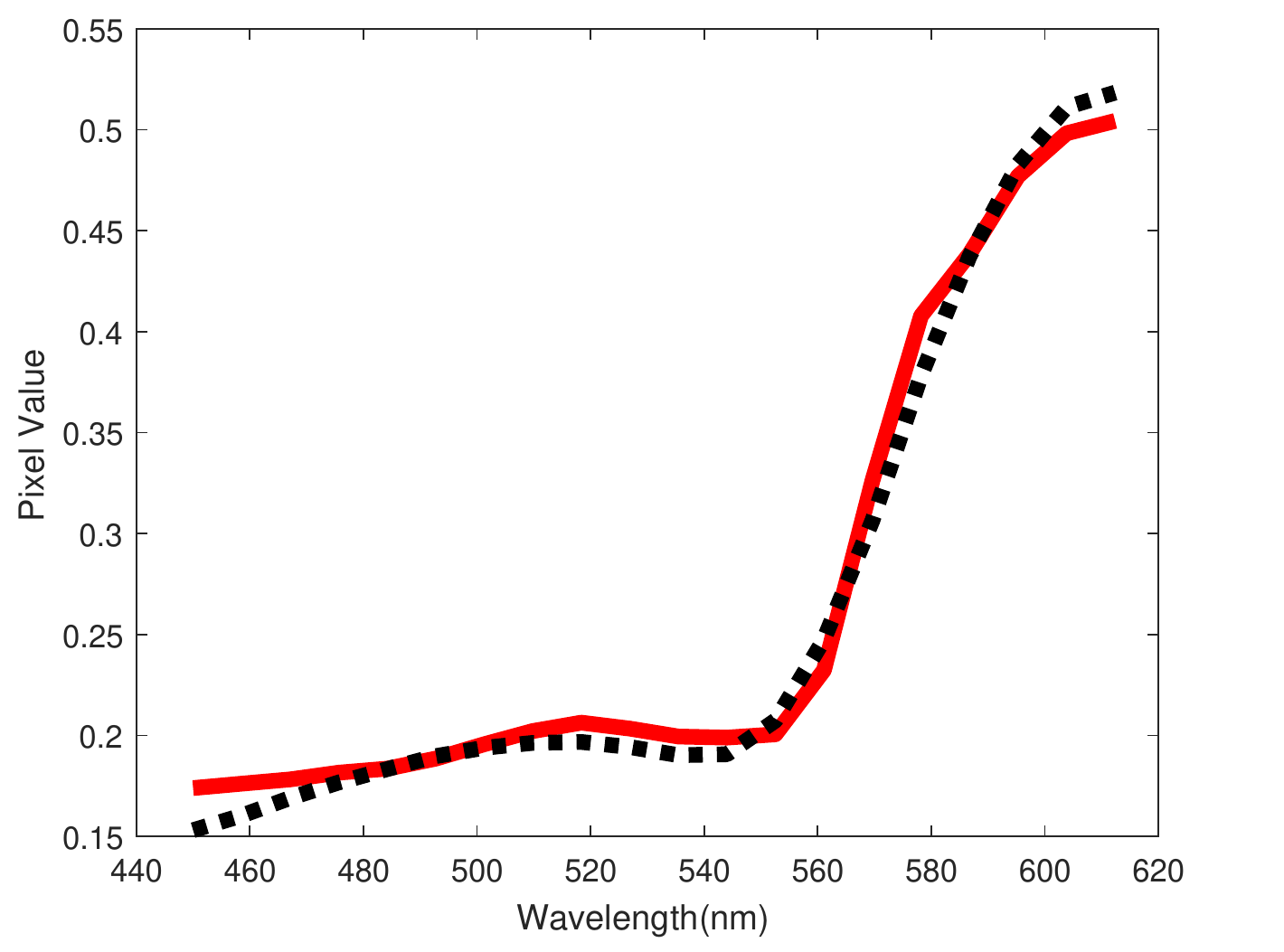}                
\end{minipage}}

\subfigure[GT]{                    
\begin{minipage}{2.25cm}
\centering                                                          
\includegraphics[width=1\linewidth]{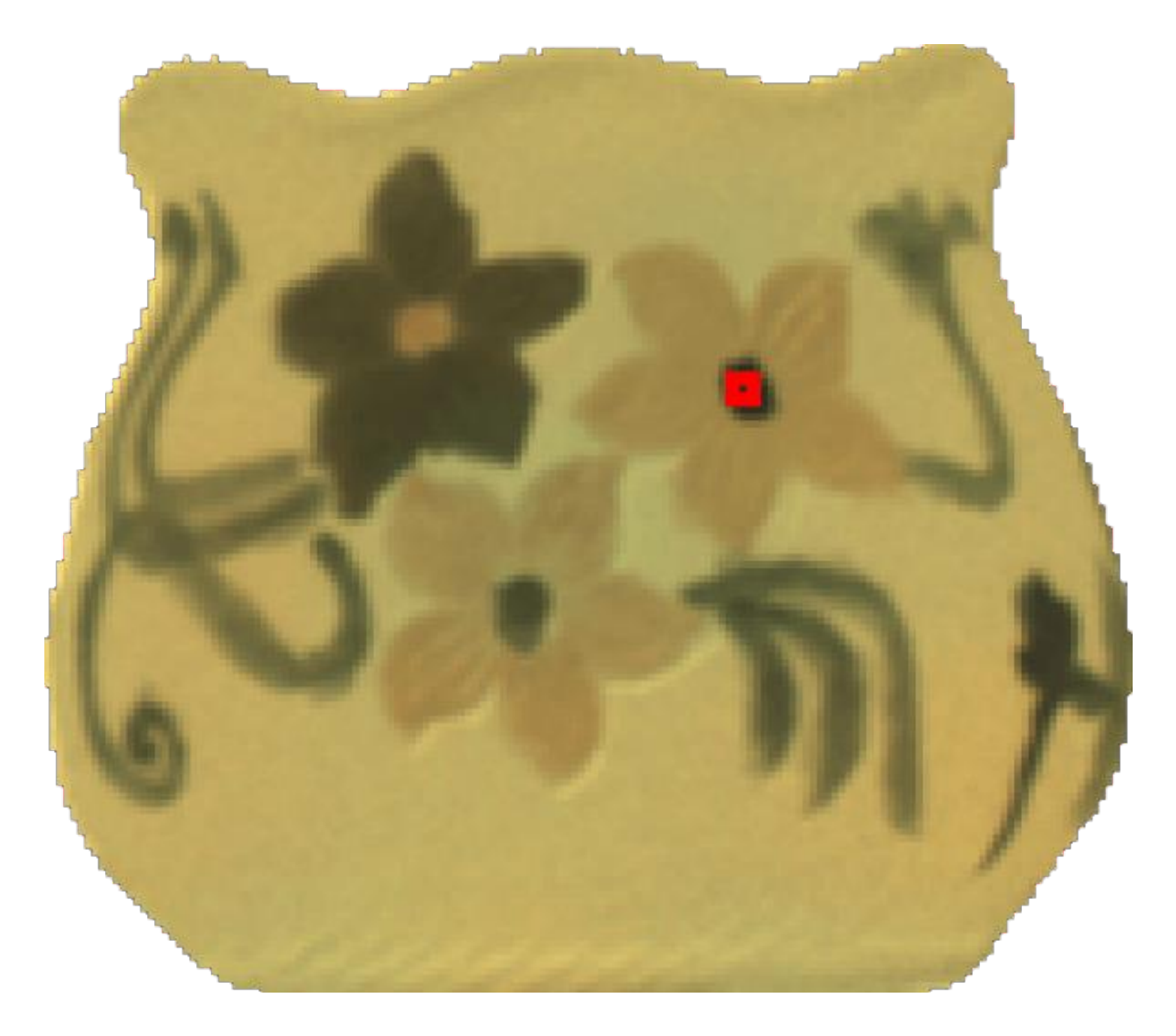}               
\end{minipage}}
\subfigure[Ours]{                    
\begin{minipage}{2.25cm}
\centering                                                          
\includegraphics[width=1\linewidth]{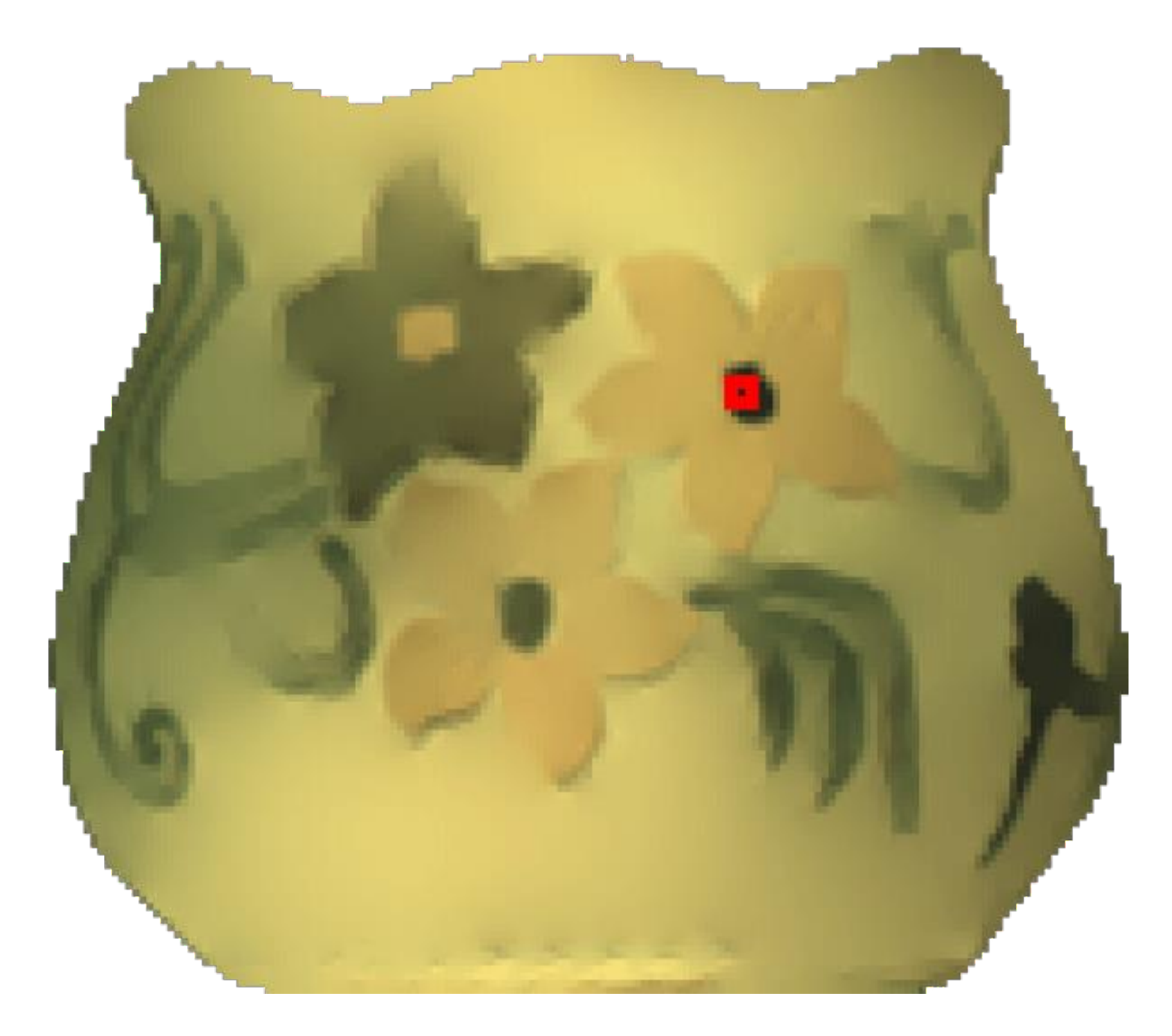}                
\end{minipage}}
\subfigure[Spectra Curve]{                    
\begin{minipage}{2.6cm}
\centering                                                          
\includegraphics[width=1\linewidth]{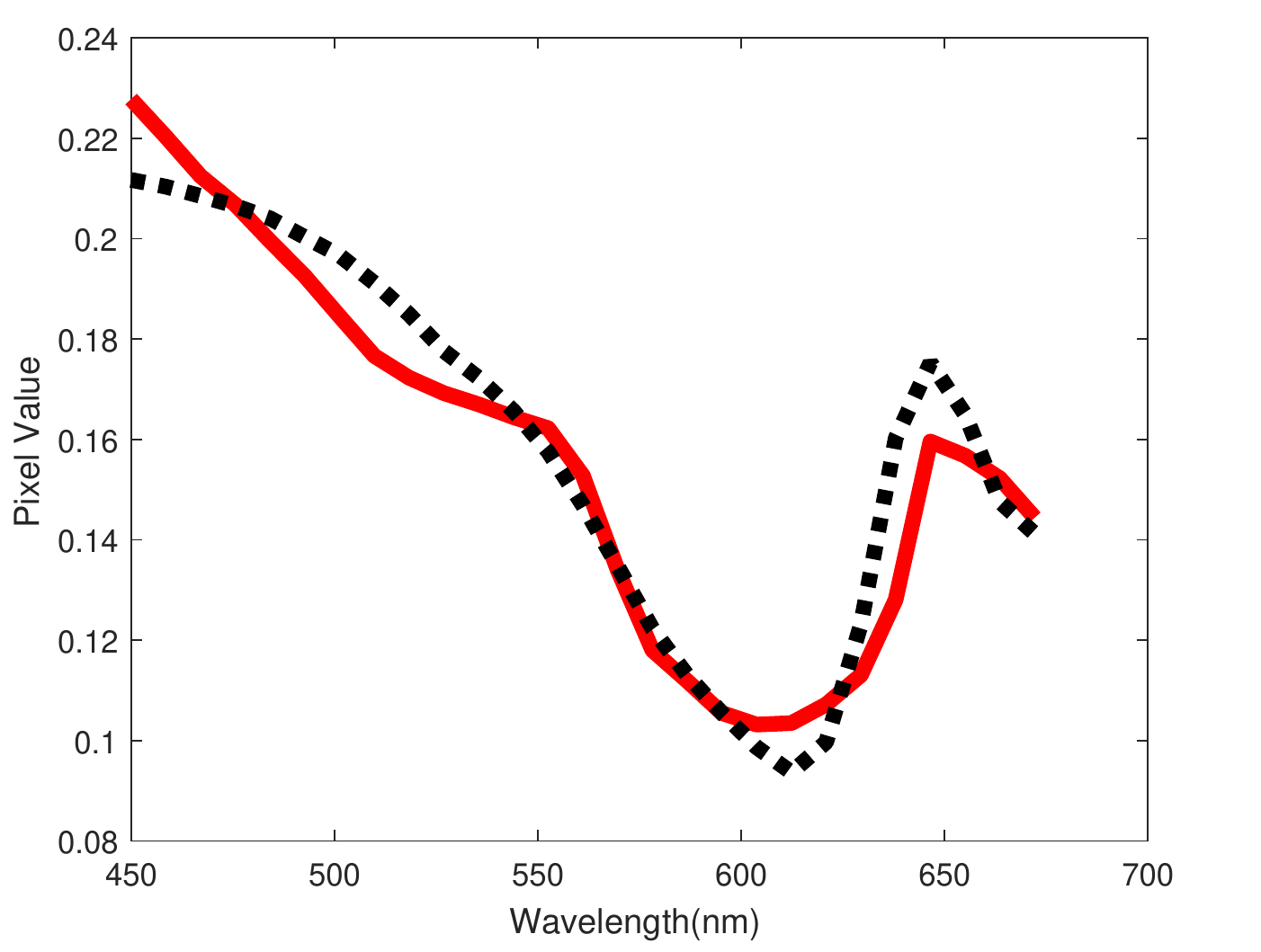}                
\end{minipage}}
\caption{(a) and (d) are ground truth of our dataset, (b) and (e) are reflectance images of our results, and the (c) and (f) are spectra curve of marked area(solid red: curve of ours; dotted black: curve of ground-truth).} 
\label{fig_curve}
\end{figure}

As to the reflectance, since it is not visualized from the spectral perspective, we try to compare the spectral curves of the reflectance from the ground truth and our algorithm. we choose patches in some scenes of our ground truth. In Fig.~\ref{fig_curve}, it is obvious that our reflectance matches well with the ground truth which means we could gain accurate spectral reflectance with better performance in computation. 

\begin{figure*}[!ht]

\centering                                                          
\subfigure{                    
\begin{minipage}{17cm}
\centering                                                          
\includegraphics[width=1\linewidth]{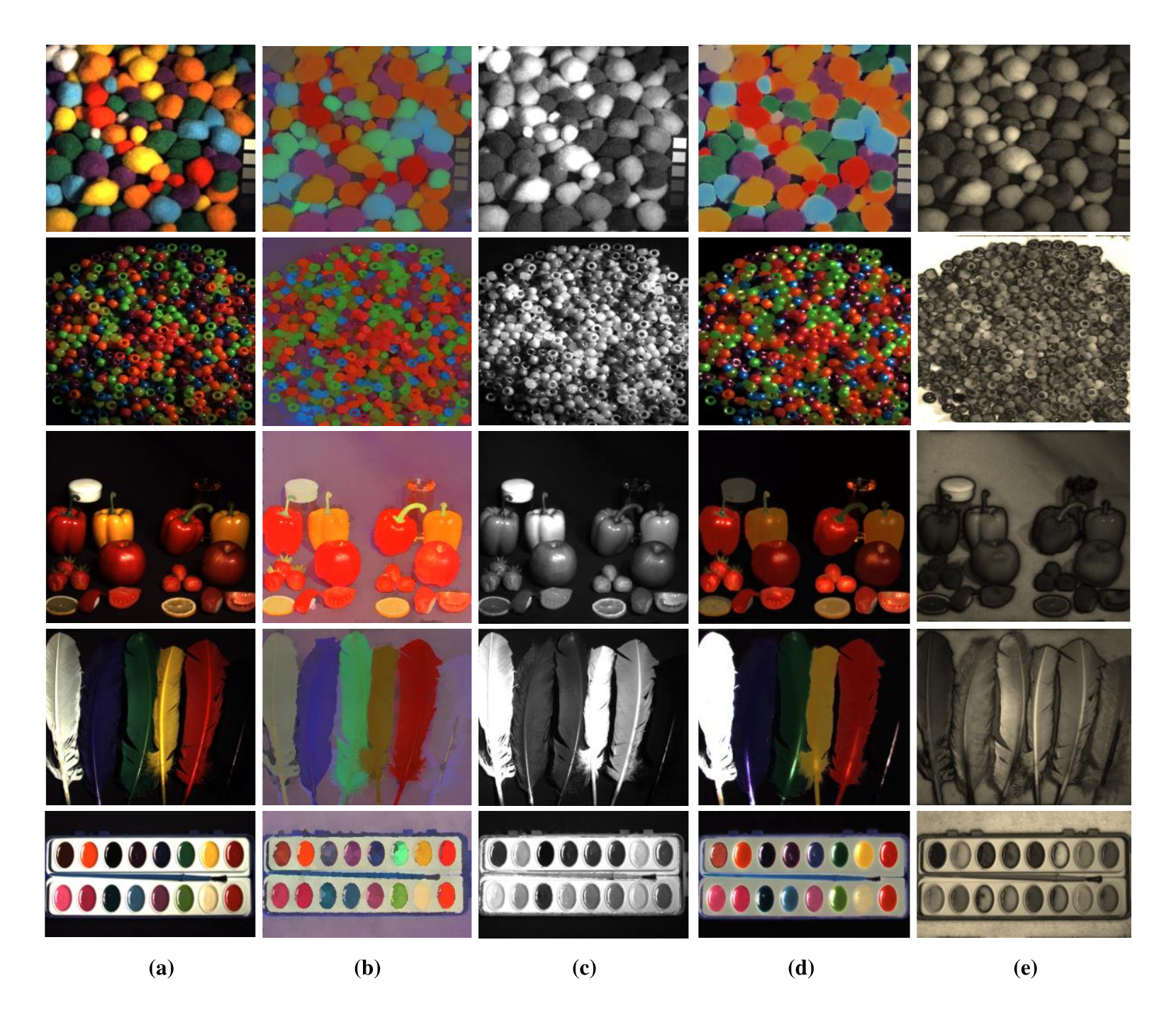}               
\end{minipage}}
\caption{Comparison of decomposition without and with the low-rank constraint. (a) A single image for five different scenarios in Nayar dataset~\cite{yasuma2010generalized}. (b)-(c) and (d)-(e) show the reflectance and shading components computed by our LRIID solution without and with low-rank constraint, respectively. Note that both the original multispectral images (a) and multispectral reflectance (d) are integrated into 3-channel images by using the response curves of RGB sensors for visualization.} 
\label{fig_nayar}
\end{figure*}

\subsection{Experiments on Nayar multispectral image database\cite{yasuma2010generalized}}
A great variety of problems assume low-dimensional subspace structure and have been solved by adding low-rank constraint, so is our method. To demonstrate the benefitd of this constraint, we compare results with and without it in Fig.~\ref{fig_nayar} using Nayar Multispectral Image Database~\cite{yasuma2010generalized}. The original images are shown in Fig.~\ref{fig_nayar}(a), the decomposed reflectance and shading without and with the low-rank constraint are shown in Fig.~\ref{fig_nayar}(b)-(e) respectively. From the comparison, we can see that the low-rank constraint helps to maintain global structures and improves decompostion results. 

\section{Conclusion}
We have addressed the problem of the recovery of reflectance and shading from a single multspectral image captured under general spectral illumination. We have applied a low rank constraint to settle the multispectral image intrinsic decomposition problem, which significantly reduced the ambiguity. gradient descent has been used to give the initial estimation of reflectance and shading, and alternating projection method has been applied to solve the bilinear problem. Experiments on our dataset have demonstrated that the performance of our work are better than prior work in multispectral domain.

Our work has left out depth information. In fact, Retinex theory fails to take effect when both shading and reflectance change extensively in local area. Shading depends on the object surface geometry, which can be derived from depth information. We hope that we can make more accurate hypothesis about shading variation using the depth and surface normal information in the future.

{\small
\bibliographystyle{ieee}
\bibliography{egbib}
}

\end{document}